\documentclass[conference]{IEEEtran}
\IEEEoverridecommandlockouts
\usepackage{cite}
\usepackage{comment}
\usepackage{amsmath,amssymb,amsfonts}
\usepackage{algorithmic}
\usepackage{graphicx}

\usepackage{amsmath}
\usepackage{textcomp}
\usepackage{xcolor}
\usepackage{algorithm}
\usepackage{algorithmic}
\def\BibTeX{{\rm B\kern-.05em{\sc i\kern-.025em b}\kern-.08em
    T\kern-.1667em\lower.7ex\hbox{E}\kern-.125emX}}

\usepackage{fancyhdr}  
\usepackage{lipsum}  

\fancypagestyle{IEEEfooter}{
    \fancyhf{}  
    \fancyfoot[C]{\small 979-8-3315-2963-5/24/\$31.00~\copyright~2024 IEEE} 
}

\begin{document}

\title{Adversarial Machine Learning: Attacking and Safeguarding Image Datasets\\

}

\author{\IEEEauthorblockN{Koushik Chowdhury}
\IEEEauthorblockA{\textit{MSc. Computer Science} \\
\textit{Saarland University}\\
Saarbrücken, Germany \\
kchy494@gmail.com}

}

\maketitle

\thispagestyle{IEEEfooter}  
\pagestyle{IEEEfooter}  

\begin{abstract}
This paper examines the vulnerabilities of convolutional neural networks (CNNs) to adversarial attacks and explores a method for their safeguarding. In this study, CNNs were implemented on four of the most common image datasets, namely CIFAR-10, ImageNet, MNIST, and Fashion-MNIST, and achieved high baseline accuracy. To assess the strength of these models, the Fast Gradient Sign Method was used, which is a type of exploit on the model that is used to bring down the models accuracies by adding a very minimal perturbation to the input image. To counter the FGSM attack, a safeguarding approach went through, which includes retraining the models on clear and pollutant or adversarial images to increase their resistance ability. The next step involves applying FGSM again, but this time to the adversarially trained models, to see how much the accuracy of the models has gone down and evaluate the effectiveness of the defense. It appears that while most level of robustness is achieved against the models after adversarial training, there are still a few losses in the performance of these models against adversarial perturbations. This work emphasizes the need to create better defenses for models deployed in real-world scenarios against adversaries.
\end{abstract}


\section{Introduction}
The field of computer vision has greatly advanced, especially through the use of deep learning, including convolutional neural networks (CNNs), and the accuracy of the field in terms of the variety of tasks assigned to image classification has been remarkable. There are also several popular image datasets, including the CIFAR-10 dataset, ImageNet, MNIST, and Fashion-MNIST, surpassing the benchmarks in these datasets, making CNNs essential for applications such as object recognition, image search engines, recommendation systems, etc. Though they are such high-tech models, it has been demonstrated that CNNs can be effectively attacked by adversarial attacks. Adversarial attacks are designed to mislead machine learning models by making small and often unnoticed alterations to the data, referred to as an input. Such situations are very alarming, in particular when the model is deployed in the areas where correct predictions are critical to safety.

Among all the adversarial attacks, one of the most powerful is the Fast Gradient Sign Method (FGSM), which creates manipulated examples by moving an input image in the direction of the model’s loss gradient \cite{b14}. This in turn suggests that even marginal changes to the input can greatly decrease the performance of the model. For example, an image that was learned as a 'cat' using CNN based on the CIFAR-10 dataset can easily be misclassified as a 'dog' after applying an FGSM attack. The same goes for models trained from scratch on ImageNet, MNIST, and Fashion MNIST, because even a small perturbation of high-resolution images or digits can cause vast misclassification.

The awareness of adversarial vulnerabilities has increased over the years, making it possible for various defense mechanisms to be produced, the most commonly known being adversarial training. Adversarial training tries to boost the model’s robustness by including adversarial examples in the training set so that the model will be able to learn from distorted inputs \cite{b14}. This process makes the model learn to generalize better over adversarial inputs, hence minimizing the impact of the attacks.

Here in this paper, an attempt is made to examine the vulnerability of CIFAR-10, ImageNet, MNIST, and Fashion-MNIST optimized CNNs to FGSM attacks. Initially, baseline accuracies for each dataset were collected before applying FGSM for generating adversarial examples; this is to determine the loss in accuracy. Then, adversarial training is performed as a defense approach wherein the models are retrained with clean as well as adversarial inputs. Lastly, the FGSM attack was conducted on the adversarially trained models to evaluate the difference with the first attack to decide the efficacy of the defense. It was noted that adversarial training generalizes model’s robustness, but they are still somewhat vulnerable, confirming the fact that adversarial defense remains an open problem.

\section{Related Work}
The Convolutional Neural Networks (CNNs) have gained high popularity when applied to images, yielding great results on the numerous datasets. This review focuses on the work done in using CNNs on the datasets like MNIST, CIFAR-10, ImageNet, and Fashion-MNIST with regards to improvement in architectures, training methodologies, and assessment metrics.

The preliminary studies by Chauhan et al.\cite{b1} have also shown the applicability of CNNs in image detection and recognition tasks and they had attained very good accuracy in MNIST as well as CIFAR-10 databases. Commencement to the investigations of CNNs, Hossain et al. \cite{b2} and Shankar et al. \cite{b3} specifically experienced with the ability of improving its performance with the correct information amount and model structure. Hossain et al. \cite{b2} noticed that, with the augmentation of a training set, the result’s accuracy was increased, and Shankar et al. \cite{b3} discussed the differences between CNN accuracy and human performance while working with the ImageNet dataset and the challenges that CNNs face in extensive and diverse tasks.

Kadam et al. \cite{b4} examined the various CNN structures and attempted to analyze how several architectures would perform in image classification with the use of MNIST and Fashion-MNIST datasets. In their experiments, they found out that the choices of activation functions, optimizers and choice of dropout affected the accuracy of the models. Also, Seng et al. \cite{b5} compared the different kinds of CNN architectures on the MNIST data set and compared them into considering figure such is the time taken for training, accuracy and size of the models.

Despite these outstanding performances, CNNs are not without flaws and they are prone to certain attacks. This basically means that it is still possible to fool CNNs and force the model to make wrong predictions, as was done by Goodfellow et al. \cite{b14}. To counter these threats, researchers have made efforts to design some defense mechanisms and they include adversarial training \cite{b7}, and robust model design \cite{b8}\cite{b9}.

Overall, the reviewed literature established the awareness that is gained on the advances that have been made to apply CNNs to image classification task on diverse datasets. Based on the previous algorithmic analysis, which process includes the initial training, adversarial attacks, and defense algorithms, remarkable progress have been made. With the help of relevant research in this field, like above, this study is expected to explore model performance, counter threats that arise from adversarial attacks, as well as attain better accuracy after defense mechanisms.

\section{Proposed Work}
The aim of this study is thus to find ways of enhancing the robustness of the CNNs in the face of adversarial attack in a more orderly manner. The proposed work includes the following step:
\begin{itemize}
\item CNNs will be trained on \textbf{four different sets of images}: CIFAR-10, ImageNet, MNIST, and Fashion MNIST. The main idea is to reach high accuracy on each dataset, setting up a ground for further experimentation.

\item Once the model has been trained successfully, \textbf{FGSM} is used to formulate the adversarial examples. This method is a form of attack in which a small perturbation to the images is made so that the distorted images would be classified in the wrong class, although they are almost the same as the original ones.

\item To \textbf{lessen the impact of FGSM}, a strategy will be added to protect the models during training by using adversarial images. This involves training on a network with clean input images and input images with adversarial perturbations. This should in turn help fill the gaps in the model so that images that have undergone the FGSM attacks are properly identified and classified.

\item In the \textbf{final step}, all safeguarded models will be tested. Metrics such as accuracy will be used to evaluate how much the testing accuracy improves with the safeguarding approach compared to when the model is attacked.
\end{itemize}

\section{Datasets}

\subsection{CIFAR-10:} This dataset called CIFAR-10 consists of 60,000 color images which were categorized to 10 classes, each image has the size of 32 x 32 pixels\cite{b10}. And thus, 50,000 images are allocated for the training and 10,000 are allocated for the test both split into 5 training batches and 1 test batch with the classes divided evenly \cite{b10}.

\subsection{ImageNet:} ImageNet \cite{b11} is an image database which contains over 100,000 synsets that are structured according to the WordNet taxonomy. There is more or less about 1,000 quality-controlled human-annotatedstill and motion image per synset to large-scale, richly annotated dataset for a spectrum of conceptions.

\subsection{MNIST:} The MNIST dataset \cite{b12} reduces 70,000 samples of handwritten digits of which 60,000 are intended for training and 10,000 for testing samples. It is famous for benchmarking/sorting machine learning algorithms on digit recognition problems. 

\subsection{Fashion-MNIST:} Fashion-MNIST \cite{b13} is the set of Zalando articles; consisting of 60,000 training images as well as 10,000 test images of gray scale having the size 28X28 pixels. The fashion MNIST is the contemporary version of the MNIST set, which is structurally as well as dimensionally similar to the MNIST, used for image classification benchmarking.

\section{Convolutional Neural Network (CNN)}
\subsection{Implementation:}
The Combined CNN Algorithm 1 regards the task of image classification as systematically focusing on different forms of image data. It commences with the preliminary preparation of the data that aims at the peculiarities of the dataset at hand. For the MNIST and Fashion-MNIST grayscale data, the image pixel values are simply divided by 255 so that they fall within a range of [0, 1]. Conversely, color datasets, for example, CIFAR-10 and ImageNet images, undergo mean normalization and scaling based on pre-determined dataset parameters.

\begin{algorithm}[H]
\caption{Combined CNN Algorithm for Image Classification}
\label{alg:combined_cnn}

\textbf{Input:}
\begin{itemize}
    \item Image data $(X, y)$ with dimensions $(n, C, H, W)$ 
    \item Number of classes $K$
    \item Hyperparameters: learning rate $\eta$, epochs $E$, batch size $B$, etc.
\end{itemize}

\textbf{Preprocessing:}
\begin{enumerate}
    \item \textbf{Normalize Image Data:}
    \begin{itemize}
        \item If $X$ is grayscale (e.g., MNIST, Fashion-MNIST):
        \[
        X' \gets \frac{X}{255}
        \]
        \item If $X$ is color (e.g., CIFAR-10, ImageNet):
        \[
        X' \gets \text{Standardize}(X)
        \]
    \end{itemize}
    
    \item \textbf{One-Hot Encode Labels:}
    \[
    y \gets \text{one-hot}(y)
    \]
\end{enumerate}

\textbf{Model Architecture:}
\begin{enumerate}
    \item \textbf{Convolutional Layer:}
    \begin{itemize}
        \item Filters: $F_1$, Size: $k_1 \times k_1$, Activation: $\sigma_1$ (e.g., ReLU)
    \end{itemize}
    
    \item \textbf{Max Pooling Layer:}
    \begin{itemize}
        \item Pooling Size: $p_1 \times p_1$
    \end{itemize}
    
    \item \textbf{Repeat Convolution and Pooling} 
    
    \item \textbf{Flatten Layer}
\end{enumerate}

\textbf{Classification Head:}
\begin{enumerate}
    \item If $K$ is small (e.g., MNIST, Fashion-MNIST):
    \begin{itemize}
        \item Dense Layer: Neurons: $N_1$, Activation: $\sigma_2$ (e.g., ReLU)
    \end{itemize}

    \item If $K$ is large (e.g., CIFAR-10, ImageNet):
    \begin{itemize}
        \item Dense Layer: Neurons: $N_1$, Activation: $\sigma_2$ (e.g., ReLU)
        \item Dropout Layer
        \item Dense Layer: Neurons: $N_2$, Activation: $\sigma_2$ (e.g., ReLU)
    \end{itemize}
    
    \item \textbf{Output Layer:}
    \begin{itemize}
        \item Neurons: $K$, Activation: softmax
    \end{itemize}
\end{enumerate}

\textbf{Training:}
\begin{enumerate}
    \item Initialize model weights randomly
    \item For each epoch $e \in \{1, \dots, E\}$:
    \begin{enumerate}
        \item Shuffle data $(X', y)$
        \item For each batch $b \in \{1, \dots, \lfloor n / B \rfloor\}$:
        \begin{enumerate}
            \item Select batch $(X_b, y_b)$
            \item \textbf{Forward Pass:} Compute activations
            \item \textbf{Backward Pass:}
            \begin{itemize}
                \item Compute loss (e.g., categorical cross-entropy)
                \item Update weights using an optimizer (e.g., Adam)
            \end{itemize}
        \end{enumerate}
    \end{enumerate}
    \item Monitor validation loss to prevent overfitting
\end{enumerate}

\textbf{Output:} Trained CNN model
\end{algorithm}

This normalization is important as it helps to achieve a uniform standard for the input data, which results in increased efficiency and convergence of a given model during the learning phase. In addition, multi-class classification is simplified by converting labels to one-hot vectors.

The structure of CNN methods has a sequential architecture with convolutional layers and pooling layers, which helps the model to identify relevant features of the input images and reduces their sizes. The number of filters optimally needed, the filter sizes, and the activation functions to be used depend on the complexity of the dataset. For some datasets with smaller class numbers, for instance, in the case of MNIST or Fashion-MNIST, the model applies very basic configurations of dense layers. However, for complex datasets with many classes in such advanced models, classically dense layers and dropout layers are added for better regularization to avoid overfitting. The softmax function was used on the last output layer as an action with regards to class probabilities to leverage the correct class from the models built.

The training of CNNs follows an iterative approach whereby the model weights are randomly initialized at the beginning and shuffling of the data is done at the beginning of each epoch in a bid to facilitate better learning. The procedure runs in the following cycle where both forward pass and backward pass are processed, in which loss is computed and weights are modified by an optimizer, for instance Adam. This is done in several cycles while the validation loss is checked so that overfitting does not manifest itself. In conclusion, this systematic approach to CNN types and their training provides good and effective ways of handling various image classifications, allowing one to learn from different datasets.

\subsection{Results:}

The convolutional neural network (CNN) achieved the following training accuracies: 84.14\% on the CIFAR-10 dataset, 81.29\% on the ImageNet dataset, 90.23\% on the MNIST, and 91.47\% on the Fashion MNIST. On testing, 78.19\% accuracy was achieved on the CIFAR-10, 74.26\% in ImageNet, 86.22\% on MNIST, and 87.27\% on Fashion MNIST. 

\begin{table}[htbp]
\caption{CNN: Train and Test Accurary for Dataset}
\begin{center}
\begin{tabular}{|c|c|c|}
\hline
\textbf{Dataset} & \textbf{Training Accuracy} & \textbf{Testing Accuracy} \\
\hline
CIFAR-10 & 84.14\% & 78.19\%\\
\hline
ImageNet & 81.29\% & 74.26\% \\
\hline
MNIST & 90.23\% & 86.22\% \\
\hline
Fashion MNIST & 91.47\% & 87.27\% \\
\hline

\end{tabular}
\label{tab1}
\end{center}
\end{table}
With relation to the CIFAR-10 dataset, the model recorded an 84.14\% accuracy during training and 78.19\% during testing. This shows that the model was well able to fit the training dataset but overfitted in the process, leading to a decline in accuracy once it was implemented into practice. Looking at the ImageNet dataset, training accuracy was 81.29\% while the test accuracy was 74.26\%. The implication and conclusion of this is that even this particular model, when performing on images of new data taken from this more complex dataset, the performance slightly dropped. 

On the other hand, with regard to the less complex MNIST dataset, as the name suggests, the model made a training accuracy of 90.23\% while test accuracy was 86.22\%. In the same way, regarding the Fashion MNIST dataset, training accuracy reported was 91.47\% while test accuracy was 87.27\%. Both the datasets exhibited a smaller gap with regards to the training and testing accuracies, indicative of good performance on new appearing data.

\section{The Fast Gradient Sign Method}
The Fast Gradient Sign Method (FGSM) is a simple method for constructing adversarial examples that was published by Goodfellow in 2014 \cite{b14}. The objective of such attacks is to generate adversarial noise that, when added to the input, will make the neural network misclassify the samples, hence greatly reducing the accuracy of the model on unforeseen data.

FGSM specifically relates to constructing adversarial examples and seeks to increase the model loss at every iteration. To obtain such adversarial examples, the inputs are first designed to decrease the model's accuracy and alter the given raw dataset. This creates some perturbed inputs that result in most distressing variations. It can be represented mathematically as follows\cite{b14}\cite{b16}:
\begin{equation}
\mathbf{x}_{\text{adv}} = \mathbf{x} + \epsilon \cdot \text{sign}(\nabla_{\mathbf{x}} J(\theta, \mathbf{x}, \mathbf{y}))  
\end{equation}

In equation 1\cite{b14}, \( \mathbf{x} \) mentioned is the seed input, and \( \epsilon \) indicates the is a small scalar value that defines how much the perturbation magnitude is to be increased, and \( \nabla_{\mathbf{x}} J(\theta, \mathbf{x}, \mathbf{y}) \) is the gradient of the loss with respect to the input\cite{b14}.

\subsection{Generating Adversarial Examples:} 
Adversarial examples were generated using the FGSM attack for all of these CNN models. Images provided as input were analyzed in terms of sensitivity by taking derivatives with respect to the input images and loss images in the backpropagation process. Otherwise, images were adjusted by a bit less than the derived perturbation and a reasonable amount of \( \epsilon \), which measures how effective this alteration is. A higher \( \epsilon \) creates a higher chance of making more drastic changes in the image, but it would make CNN less effective.

\subsection{Perturbation Magnitude (\( \epsilon \)):} The phenomenon of transferring the distance between action choices on the down side in response to the questions where \( \epsilon \) has its global value and motion affects its local value. A moderate Epsilon value (\( \epsilon \) = 0.1) was used to decrease \( \epsilon \) and make it as low as possible without omitting the effect.

\subsection{Effect on Test Accuracy:}
Each of these CNN models was run again using adversarial inputs mentioned above. Test accuracy for all the models was severely impacted due to the FGSM attack. The following Table II provides the original test accuracy and the respective reduced test accuracy post-FGSM application with (\( \epsilon \) = 0.1).
\begin{table}[htbp]
\caption{Effect on Test Accuracy after FGSM }
\begin{center}
\begin{tabular}{|c|c|c| c|}
\hline
\textbf{Dataset} & \textbf{Original} & \textbf{After FGSM} & \textbf{Reduction}\\
\hline
CIFAR-10  & 78.19\% & 49.32\% & 28.87\%\\
\hline
ImageNet  & 74.26\% & 46.70\% & 27.56\%\\
\hline
MNIST  &86.22\% & 65.87\% & 20.35\%\\
\hline
Fashion MNIST  & 87.27\% & 60.45\% & 26.82\% \\
\hline
\end{tabular}
\label{tab1}
\end{center}
\end{table}
However, once the FGSM attack was made by using an \( \epsilon \) value of 0.1, nothing less significant from the fact that the test accuracies of four CNN models registered a significant reduction. For the CIFAR-10 dataset, the accuracy dropped from a stunning 78.19\% to 49.32\%, indicating how hard this model was affected. Similarly, the ImageNet accuracy reduced from 74.26\% to 46.70\%. The accuracy of the MNIST model was cut down from 86.22\% to 65.87\%, signifying the extent of the vulnerability of the model to adversarial perturbations. The last model, Fashion MNIST, dropped in accuracy growth from 87.27\% down to 60.45\%, indicating the decline in its performance consequent to the attack.
\subsection{Sample Examples After FGSM Attack:}
\subsubsection{Example From CIFAR-10}
Fig. 1 presents two images side by side with the purpose of demonstrating the impact of the FGSM attacks after the CNN. The left image, titled ‘Before FGSM Attack’ shows the image with the correct class label as a truck, and it predicts correctly. 
\begin{figure}[htbp]
\centerline{\includegraphics[width=0.47\textwidth]{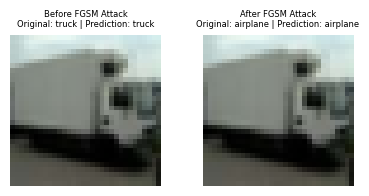}}
\caption{CIFAR-10: Effect after CNN and FGSM}
\label{fig}
\end{figure}
The right image, referred to as 'After FGSM Attack', displays the image after the application of the Fast Gradient Sign Method (FGSM) attack with the new prediction (an airplane), which is a wrong prediction. This visualization more or less shows how adversarial perturbations affect the CNN to give certain outputs and is a good illustration of the effect of these attacks on accuracy.

\subsubsection{Example From Fashion MNIST}
In Fig. 2,  ‘Original’ refers to the actual image of the assigned label, and prediction refers to what the model predicted. 
\begin{figure}[htbp]
\centerline{\includegraphics[width=0.44\textwidth]{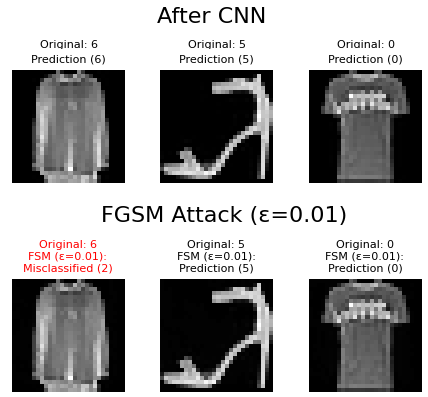}}
\caption{Fashion MNIST: Effect after CNN and FGSM}
\label{fig}
\end{figure}
Take the first image, for example, which is a ‘6’ according to the model’s prediction; the model has correctly understood the label. 

After the FGSM attack, the image has a little change and because of this, the prediction of the model altered, and it replaced the image with another-looking object having a label that does not in any way correlate with the original image.

This means that the FGSM attack has been able to cause some form of impairment to the model so that after applying the attack, the model has misclassified the image even though the model was able to correctly identify the label when the image was not attacked in any way. This shows how the attack has managed to succeed in degrading the model's capability to classify the image accurately.

\section{Safeguarding}
Thus, in order to prevent the effects of FGSM and other attacks on the performance of models, adversarial training and other solutions should be used. Implementation of adversarial training is the focus of discussion in the next subsection.

\subsection{Adversarial Training}
It is equally important to understand that among the widely used attacks, FGSM attacks can be prevented through the use of adversarial training. The following Algorithm~\ref{alg:adversarial_training} is developed based on the work done by Goodfellow et al.\cite{b14} on exploit adversarial examples.
\begin{algorithm}
\caption{Adversarial Training for FGSM Defense}
\label{alg:adversarial_training}
\begin{algorithmic}[1]
\STATE \textbf{Input:} Model $f_{\theta}$, training dataset $\mathcal{D}$, loss function $\mathcal{L}$, perturbation magnitude $\epsilon$
\STATE \textbf{Output:} Trained adversarially robust model $f_{\theta}$

\STATE Initialize model parameters $\theta$
\FOR{each minibatch $(x, y) \in \mathcal{D}$}
    \STATE Determine the gradient of the loss \cite{b16} \cite{b6}:
    \[
    \nabla_x \mathcal{L}(f_{\theta}(x), y)
    \]
    \STATE Generate adversarial example using FGSM \cite{b14}\cite{b16}:
    \[
    x_{\text{adv}} = x + \epsilon \cdot \text{sign}(\nabla_x \mathcal{L}(f_{\theta}(x), y))
    \]
    \STATE Combine original and adversarial examples into a new minibatch:
    \[
    \tilde{x} = [x, x_{\text{adv}}], \quad \tilde{y} = [y, y]
    \]
    \STATE Compute the model prediction on the new minibatch:
    \[
    \hat{y} = f_{\theta}(\tilde{x})
    \]
    \STATE Update the model parameters using gradient descent:
    \[
    \theta = \theta - \eta \cdot \nabla_{\theta} \mathcal{L}(f_{\theta}(\tilde{x}), \tilde{y})
    \]
\ENDFOR
\STATE \textbf{Return:} Trained model $f_{\theta}$
\end{algorithmic}
\end{algorithm}

The authors, Goodfellow et al. \cite{b14}, also detailed how to produce adversarial images through FGSM. This is outlined in line 6 of the representation of Algorithm~\ref{alg:adversarial_training}. The original image is then summed with the grade of the loss to get an adversarial example. The value of (\( \epsilon \)) is modified by the models that have been used for particular datasets. In the 7th line of the Algorithm~\ref{alg:adversarial_training}, the original data and the adversarial examples are concatenated side by side to form a new set of data that will be imposed for training the system. However, the model prediction is calulated, and this time the minibatch is used, which is a combination obtained from line 7 of the Algorithm~\ref{alg:adversarial_training}. Consequently, for each epoch, the weights of the model were adjusted in terms of the estimated loss over the augmented dataset to reach an optimization point. The following Table III shows how the test accuracy improves when safeguarding is employed while attacking with FGSM. Without the use of adversarial training, accuracy decreases significantly under attack; however, with the use of adversarial training, the model has better protection, with accuracy significantly improving for all the datasets, particularly and most obviously for the simpler datasets such as the MNIST and Fashion MNIST. But the advantage is even less pronounced for more complicated data sets such as CIFAR-10 and ImageNet.
\begin{table}[htbp]
\caption{FGSM Effect on Test Accuracy on Initial-Trained and Adversarially Trained}
\begin{center}
\begin{tabular}{|c|c|c| c|}
\hline
\textbf{Dataset} & \textbf{\( \epsilon \)} & \textbf{Initial} & \textbf{Adversarially Trained}\\
\hline
CIFAR-10  & 0.03 & 49.32\% &71.17\%\\
\hline
ImageNet  & 0.03 & 46.70\% & 70.01\%\\
\hline
MNIST  &0.1 & 65.87\% & 80.25\%\\
\hline
Fashion MNIST  & 0.1 & 60.45\% & 81.82\% \\
\hline
\end{tabular}
\label{tab1}
\end{center}
\end{table}

\subsection{Final Thoughts}

Fig. 3 depicts all the datasets and how they are vulnerable to adversarial attacks. On the x-axis, 4 datasets are labeled, namely CIFAR-10, ImageNet, MNIST, and Fashion MNIST, while the y-axis plots the accuracy of a model on each of the databases. The fig. 3 shows that when FGSM is applied to the image datasets, the accuracy level of the models deteriorates considerably for all the datasets. Also, as shown in Fig. 3, adversarial training can improve model robustness against these attacks, leading to higher accuracy compared to models trained without adversarial examples.

\begin{figure}[htbp]
\centerline{\includegraphics[width=0.5\textwidth]{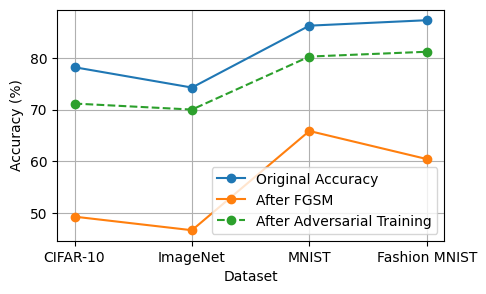}}
\caption{Accuracy Comparison Before and After FGSM and Adversarial Training}
\label{fig}
\end{figure}

Initially, the testing accuracy is more than 74\% for both the CIFAR-10 and ImageNet datasets [Table I], but the model performed exceptionally well for the MNIST and Fashion MNIST datasets, getting testing accuracy more than 85\% for both. Also for both MNIST and Fashion MNIST, the reduction of accuracy after FGSM attack is less than for both CIFAR-10 and ImageNet. 

It is noticeable that both CIFAR-10 and ImageNet drop highly, respectively, 28.87\% and 27.56\% percent [Table II]. But after the adversrial training, the accuracy drop for all datasets after the attack is comparatively less. The result can also be improved more with complex architecture, hyperparameter tuning, regulkarization techniques, etc.

\section{Conclusion}
In this research, CNNs were applied to four widely used datasets: CIFAR-10, ImageNet, MNIST, and Fashion MNIST datasets and tested the robustness against adversarial attacks. First, it was noticed that the model performs with high accuracy for all the datasets and is able to classify the images well in the normal circumstances, where there is no adversarial tampering. However, the model showed poor performance during the conventional attack, such as Fast Gradient Sign Method (FGSM), as its accuracy was decreased for CIFAR-10 and ImageNet by the adversarial perturbation. MNIST as well as Fashion MNIST appeared to have a slightly reduced accuracy, which suggests that the features of this set of image databases were less sensitive to attacks of such kinds. To this end, the adversarial training approach was adopted, which is an attempt to boost the model’s ability to resist adversarial attacks. After this process, it is noticeable that the accuracy of the model increased significantly for all datasets, but it is compared with the beginning CIFAR-10 and ImageNet, especially. This result secures the finding of prior work that adversarial training significantly improves model robustness at the expense of a minor loss in classification accuracy with adversarial inputs. In total, adversarial training decreases the FGSM attack effect and allows the models to have greater accuracy in adversarial settings.

There are some suggestions where the machine learning models could be made more robust to the adversarial perturbations in future research. First, it would be advantageous to extend the adversarial training scheme as regards to a number of the attack techniques and their perturbation methods. Another idea could be seeking mixed defense methods, where adversarial training is complemented with alternative approaches such as defensive distillation or robust optimization. In addition, the research community will need to allow for the consolidation of research by interacting with others to share the results and the methods. Working on developing open-source tools to enhance the adversarial robustness of machine learning systems has the potential of speeding up progress in this area. Last but not least, it will be helpful to enhance the interaction with the research sphere by reporting the results of the developed approach and the ongoing studies to enhance open-source tools and methods for adversarial robustness.

\vspace{12pt}
\color{red}

\end{document}